\documentclass[10pt,twocolumn,letterpaper]{article}

\usepackage[pagenumbers]{cvpr}

\usepackage{graphicx}
\usepackage{amsmath}
\usepackage{amssymb}
\usepackage{booktabs}

\usepackage[accsupp]{axessibility}

\usepackage[pagebackref,breaklinks,colorlinks]{hyperref}

\usepackage[capitalize]{cleveref}
\crefname{section}{Sec.}{Secs.}
\Crefname{section}{Section}{Sections}
\Crefname{table}{Table}{Tables}
\crefname{table}{Tab.}{Tabs.}

\def\cvprPaperID{9}
\def\confName{CVPR}
\def\confYear{2023}

\begin{document}

\title{Fast GraspNeXt: A Fast Self-Attention Neural Network Architecture for Multi-task Learning in Computer Vision Tasks for Robotic Grasping on the Edge}

\author{Alexander Wong$^{1,2,3}$ \qquad Yifan Wu$^{1}$ \qquad Saad Abbasi$^{1,3}$ \qquad Saeejith Nair$^{1}$ \\ Yuhao Chen$^{1}$ \qquad Mohammad Javad Shafiee$^{1,2,3}$\\
			$^{1}$ Vision and Image Processing Research Group, University of Waterloo\\
			$^{2}$ Waterloo Artificial Intelligence Institute, Waterloo, ON\\
			$^{3}$ DarwinAI Corp., Waterloo, ON\\ 
{\tt\small {\{a28wong, yifan.wu1, srabbasi, smnair, yuhao.chen1, mjshafiee\}}@uwaterloo.ca}
}
\maketitle

\begin{abstract}
Multi-task learning has shown considerable promise for improving the performance of deep learning-driven vision systems for the purpose of robotic grasping.  However, high architectural and computational complexity can result in poor suitability for deployment on embedded devices that are typically leveraged in robotic arms for real-world manufacturing and warehouse environments.  As such, the design of highly efficient multi-task deep neural network architectures tailored for computer vision tasks for robotic grasping on the edge is highly desired for widespread adoption in manufacturing environments.  Motivated by this, we propose Fast GraspNeXt, a fast self-attention neural network architecture tailored for embedded multi-task learning in computer vision tasks for robotic grasping.  To build Fast GraspNeXt, we leverage a generative network architecture search strategy with a set of architectural constraints customized to achieve a strong balance between multi-task learning performance and embedded inference efficiency.  Experimental results on the MetaGraspNet benchmark dataset show that the  Fast GraspNeXt network design achieves the highest performance (average precision (AP), accuracy, and mean squared error (MSE)) across multiple computer vision tasks when compared to other efficient multi-task network architecture designs, while having only 17.8M parameters (about $>$5$\times$ smaller), 259 GFLOPs (as much as $>$5$\times$ lower) and as much as $>$3.15$\times$ faster on a NVIDIA Jetson TX2 embedded processor.

\end{abstract}

\begin{figure}[t]
    \centering
    \begin{subfigure}[t]{0.48\linewidth}
        \includegraphics[width=\textwidth]{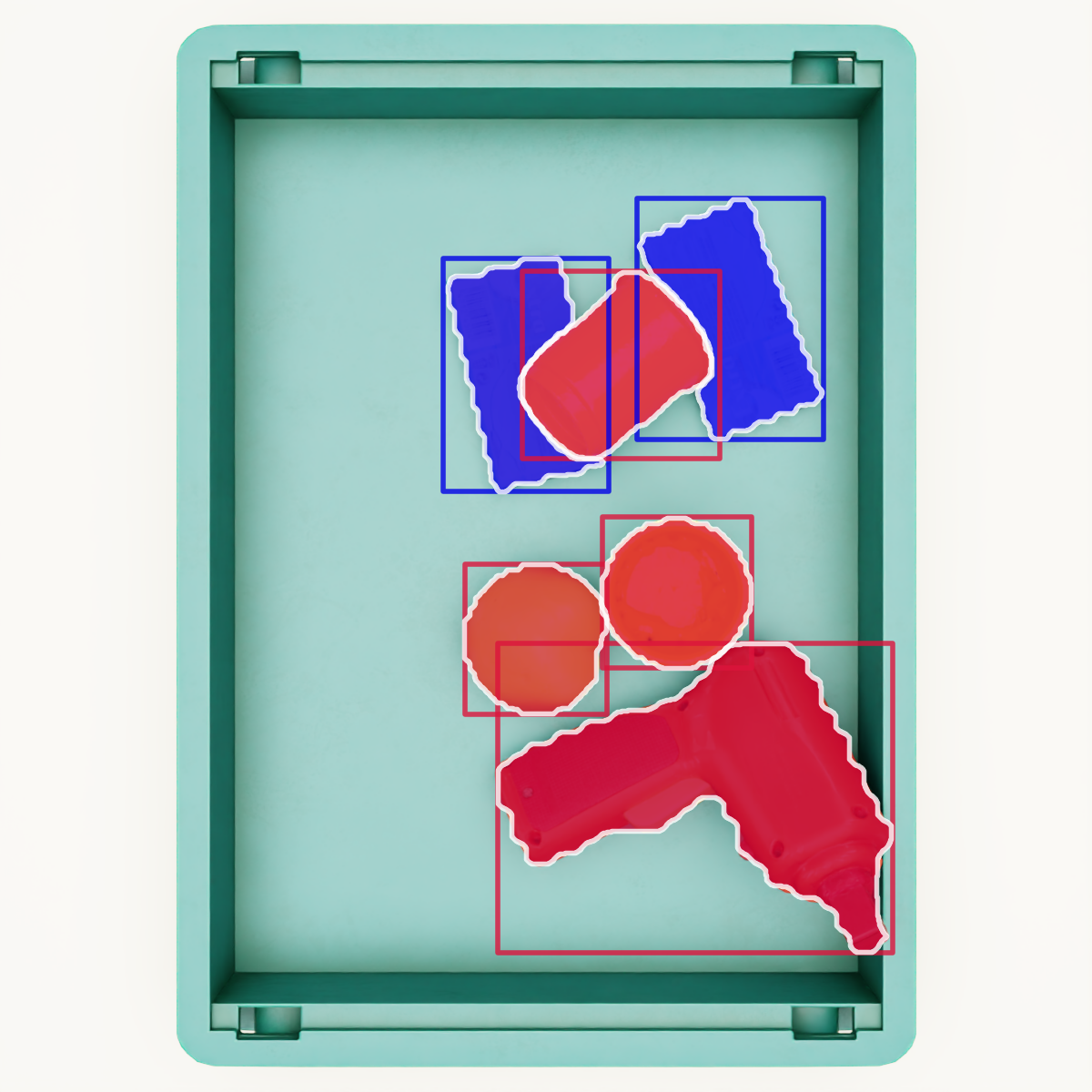}
        \caption{Visible mask.}
    \end{subfigure}
    \hfill
    \begin{subfigure}[t]{0.48\linewidth}
        \includegraphics[width=\textwidth]{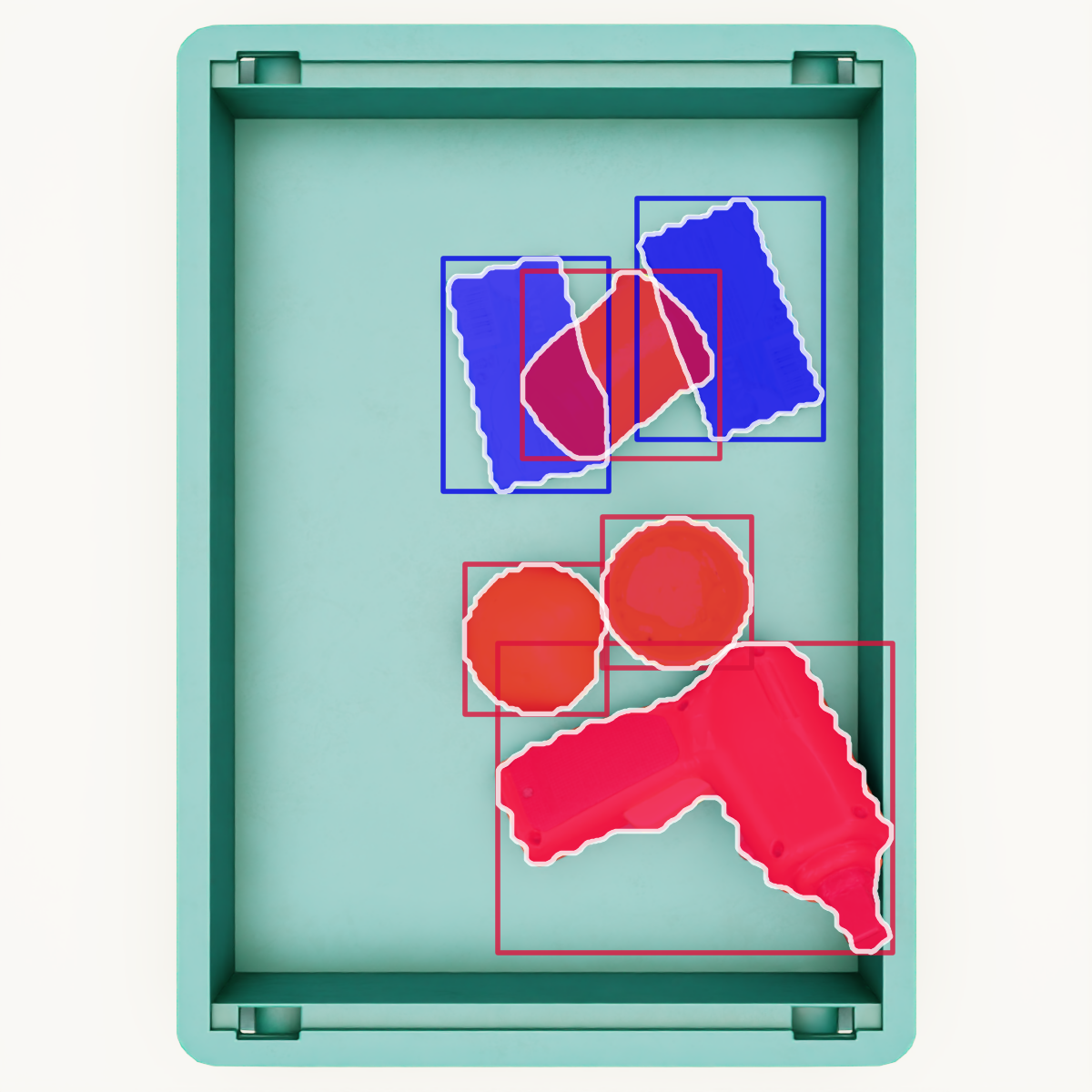}
        \caption{Amodal mask.}
    \end{subfigure}
    \begin{subfigure}[t]{0.48\linewidth}
        \includegraphics[width=\textwidth]{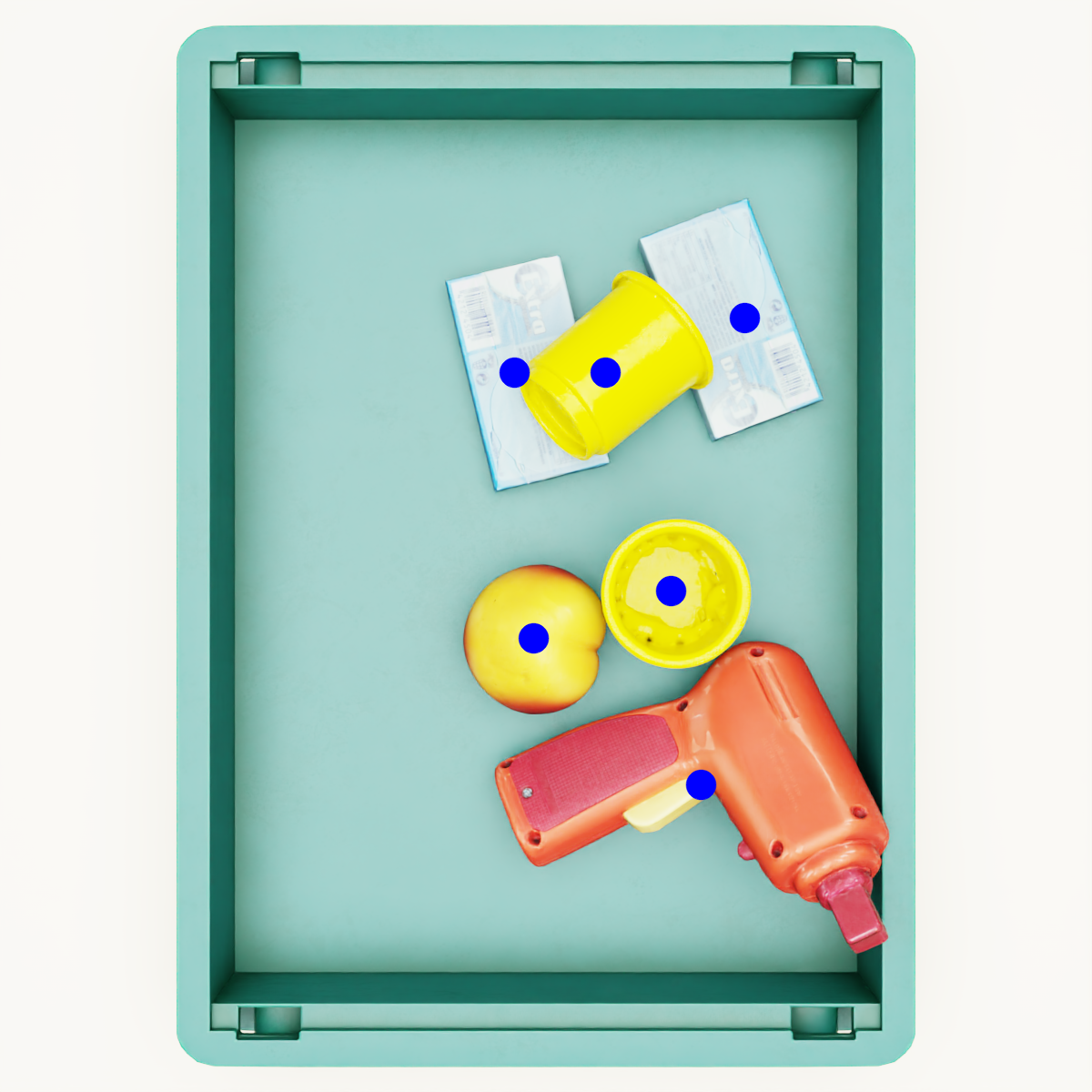}
        \caption{Center of mass.}
    \end{subfigure}
    \hfill
    \begin{subfigure}[t]{0.48\linewidth}
        \includegraphics[width=\textwidth]{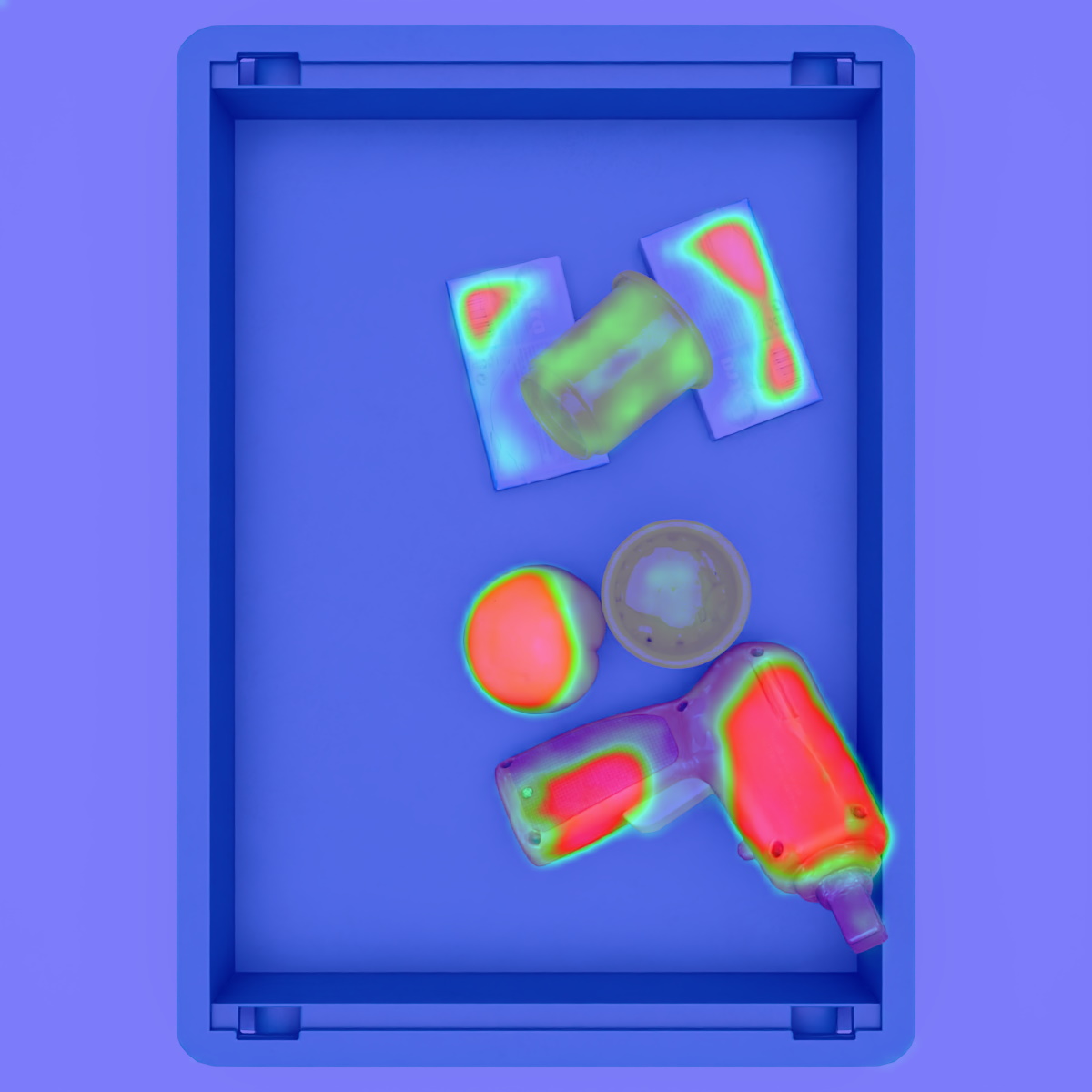}
        \caption{Suction grasp heatmap.} 
    \end{subfigure}
    \caption{Example multi-task outputs from Fast GraspNeXt. (a) and (b) Detected occluded objects are shown in blue and non-occluded objects are shown in red. (c) The detected center of mass of each object is shown in blue. (d) Applicability of suction grasp is labelled from high to low in red, green, and blue as a heatmap.}
    \label{fig:example}
\end{figure}

\section{Introduction}
\label{sec:intro}

\begin{figure*}[t]
    \includegraphics[width=\textwidth]{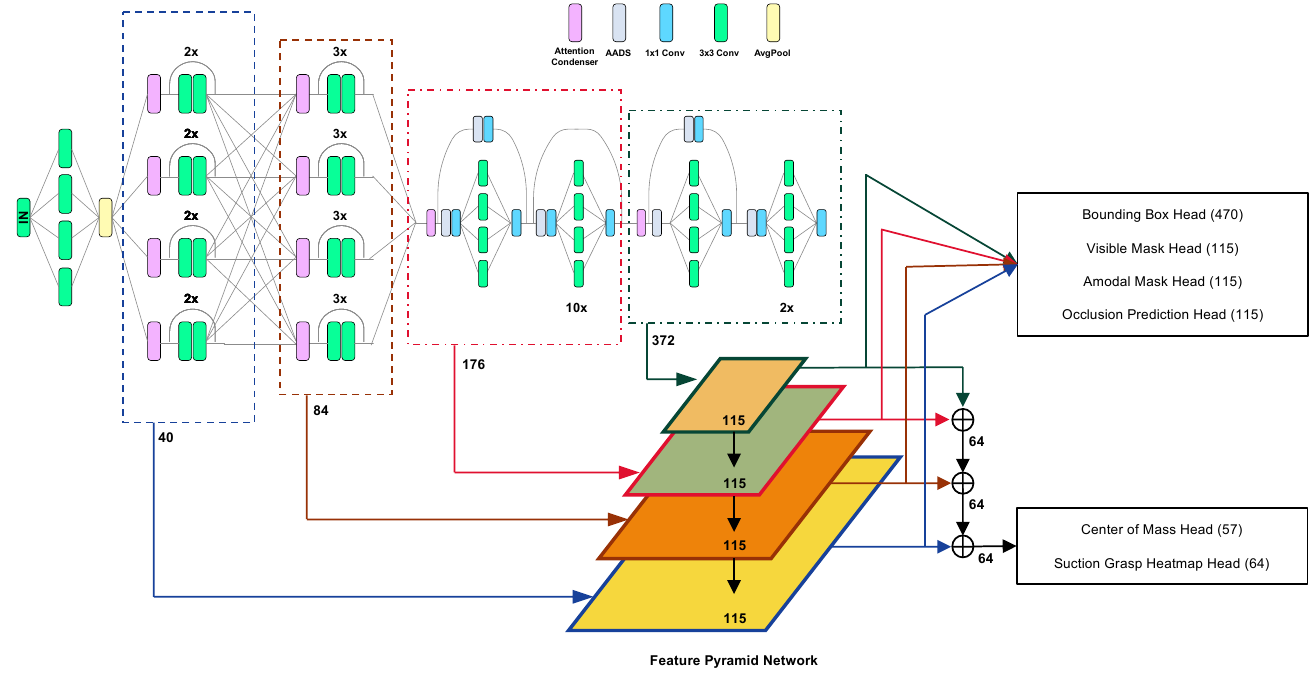}
\caption{Overall network architecture design for Fast GraspNeXt, which possess a self-attention neural network architecture with highly optimized macroarchitecture and microarchitecture designs for all components. Fast GraspNeXt consists of a generated self-attention backbone architecture feeding into a generated feature pyramid network architecture followed by generated head network architecture designs for multi-task learning. The numbers in brackets are channel sizes of the feature maps in the heads.}
    \label{fig:architecture}
\end{figure*}
Significant advances have been made in recent years to take advantage of deep neural networks for robotic grasping.  In particular, multi-task learning has shown considerable promise for improving the performance of deep learning-driven vision systems for robotic grasping~\cite{dexnet, prew2021improving,DUAN2023106059}, where the underlying goal is to learn to perform additional tasks during the model training process. Multi-task learning has enabled not only greater precision and versatility in deep learning-driven vision systems for robotic grasping, but also enabled such systems to perform a wide range of computer vision tasks that are important for robotic grasping (see \cref{fig:example} for example tasks that need to be performed by such deep learning-driven vision systems for robotic grasping such as visible object mask detection, amodal object detection \cite{uoais}, center of mass prediction, and suction grasp heatmap generation \cite{suctionnet1b}). However, while multi-task learning can greatly improve the performance of computer vision tasks for robotic grasping, high architectural and computational complexity can limit operational use in real-world manufacturing and warehouse environments on embedded devices.

Motivated to address these challenges with embedded deployment for robotic grasping in real-world manufacturing and supply chain environments, we  leverage a generative network architecture search strategy with a set of architectural design constraints defined to achieve a strong balance between multi-task learning performance and embedded operational efficiency.  The result of this generative network architecture search approach is Fast GraspNeXt, a fast self-attention neural network architecture tailored specifically for multi-task learning in robotic grasping under embedded scenarios.  

The paper is organized as follows.  Section 2 describes the methodology behind the creation of the proposed Fast GraspNeXt via generative network architecture search, as well as a description of the resulting deep neural network architecture.  Section 3 describes the dataset used in this study, the training and testing setup, as well as the experimental results and complexity comparisons.

\section{Methods}

\subsection{Generative Network Architecture Search}

In this paper, we take a generative network architecture search approach to creating the optimal multi-task deep neural network architecture for Fast GraspNeXt.  More specifically, we leveraged the concept of generative synthesis~\cite{wong2018ferminets}, an iterative method that generates highly tailored architectural designs that satisfy given requirements and constraints (e.g., model performance targets). Generative synthesis can be formulated as a constrained optimization problem:
\begin{equation}
\mathcal{G}=\max_{\mathcal{G}}\mathcal{U}(\mathcal{G}(s)) \;\;\;\textrm{ subject to} \;\;\; 1_r(G(s))=1, \;\;\forall\in \mathcal{S}.
\end{equation}
\noindent where the underlying objective is to learn an expression $\mathcal{G}(\cdot)$ that, given seeds $\{s|s \in S\}$, can generate network architectures $\{N_s|s \in S\}$ that maximizes a universal performance metric $U$ (e.g.,~\cite{wong2019netscore}) while adhering to operational constraints set by the indicator function $1_r(\cdot)$. This constrained optimization is solved iteratively through a collaboration between a generator $G$ and an inquisitor $I$ which inspects the generated network architectures and guides the generator to improve its generation performance towards operational requirements (see~\cite{wong2018ferminets} for details).

To build Fast GraspNeXt, we enforce essential design constraints through $1_r$(·) in Eq.~1 to achieve the desired balance between i) accuracy, ii) architectural complexity, and iii) computational complexity to yield high-performance, compact, and low-footprint neural network architectures such as:
\begin{enumerate}
\item Encouraging the implementation of anti-aliased downsampling (AADS)~\cite{zhang2019shiftinvar} to enhance network stability and robustness.
\item Encouraging the use of attention condensers~\cite{wong2020tinyspeech}, which are highly efficient self-attention mechanisms designed to learn condensed embeddings characterizing joint local and cross-channel activation relationships for selective attention.  They have been shown to improve representational performance while improving efficiency at the same time.
\item Enforce a FLOPs requirement of less than 300B FLOPs and an accuracy requirement of no lower AP across all assessable tasks than a ResNet-50 variant of the multi-task network for robotic grasping (which we call ResNet-GraspNeXt) by 0.5\%.
\end{enumerate}

\subsection{Network Architecture}
The resulting Fast GraspNeXt network architecture design is shown in~\cref{fig:architecture}.  It possesses a self-attention neural network architecture with highly optimized macroarchitecture and microarchitecture designs for all its components. The network architecture adheres to the  constraints we imposed, with the generated backbone architecture feeding into a generated feature pyramid network architecture design followed by generated head network architecture designs for predicting the multi-task outputs: i) amodal object bounding boxes, ii) visible object masks, iii) amodal object masks, iv) occlusion predictions, v) object center of mass, vi) and suction grasp heatmap. 

More specifically, the multi-scale features from the generated backbone architecture are provided as input directly to each level of the generated feature pyramid network architecture, followed by the generated bounding box head, visible mask head, amodal mask head and occlusion prediction head. Each level of the feature pyramid network are also upsampled to reach the same scale and summed as input for the center of mass head and suction grasp heatmap head.  

The multi-task training loss, denoted as $L_{mt}$, used to train Fast GraspNeXt is a weighted combination of task-specific losses and can be expressed by
\begin{equation}
\begin{split}
    L_{mt} &= {l_{rpn}+\lambda_{1}l_{abox}+\lambda_{2}l_{segm\_v}+\lambda_{3}l_{segm\_a}} \\ 
    &+\lambda_{4}l_{occ}+\lambda_{5}l_{com}+\lambda_{6}l_{suc}
\end{split}
\end{equation}
\noindent where $\lambda_1, \lambda_2, \ldots, \lambda_6$ denote task-specific weight coefficients used to balance the contribution of individual task-specific losses. The individual task-specific losses are defined as follows:
\begin{itemize}
\setlength\itemsep{0.00em}

    \item $l_{rpn}$: Region Proposal Network loss \cite{fasterrcnn}
    \item $l_{abox}$: Amodal bounding box prediction loss \cite{uoais}
    \item $l_{segm\_v}$: Visible mask segmentation loss \cite{uoais}
    \item $l_{segm\_a}$: Amodal mask segmentation loss \cite{uoais}
    \item $l_{occ}$: Occlusion classification loss \cite{uoais}
    \item $l_{com}$: Center of mass heatmap prediction loss implemented with the modified focal loss proposed by CenterNet \cite{centernet}
    \item $l_{suc}$: Suction grasp heatmap prediction loss implemented with pixel-wise averaged mean squared error (MSE) loss
    
\end{itemize}

It can be observed that the architecture design is highly heterogeneous and columnar for high architectural and computational efficiency.  It can also be observed that the architecture design possesses attention condensers at different stages of the architecture for improved attentional efficacy and efficiency.  Furthermore, the architecture design possesses AADS at strategic locations for greater robustness.  Finally, it can be observed that the macroarchitecture for each task-specific head is unique, thus tailored around the specific balance between accuracy and efficiency for each individual task.  As such, these characteristics make the Fast GraspNeXt architecture design well-suited for high-performance yet highly efficient multi-task robotic grasp applications on the edge.

\section{Experiments}

\begin{table*}[t]
\centering
\caption{Summary of quantitative performance results on MetaGraspNet dataset and network complexity.}
\label{tab:results}
\resizebox{\textwidth}{!}{
\begin{tabular}{lccccccccc}
\toprule
Model                  & \begin{tabular}[c]{@{}c@{}}Inf. Time\\ (ms)\end{tabular} & \begin{tabular}[c]{@{}c@{}}Amodal\\ Bbox AP\end{tabular} & \begin{tabular}[c]{@{}c@{}}Visible\\ Mask AP\end{tabular} & \begin{tabular}[c]{@{}c@{}}Amodal\\ Mask AP\end{tabular} & \begin{tabular}[c]{@{}c@{}}Occlusion\\ Accuracy\end{tabular} & \begin{tabular}[c]{@{}c@{}}Center of\\ Mass AP\end{tabular} & \begin{tabular}[c]{@{}c@{}}Heatmap\\ MSE\end{tabular} & \begin{tabular}[c]{@{}c@{}}Parameters\\ (M)\end{tabular} & \begin{tabular}[c]{@{}c@{}}FLOPs\\ (B)\end{tabular} \\ \midrule
ResNet-GraspNeXt       & 3501                                                     & 85.0\%                                                   & 84.9\%                                                    & 84.1\%                                                   & 77.2\%                                                       & 75.3\%                                                      & 0.0113                                                & 92.1                                                     & 1314                                                \\
EfficientNet-GraspNeXt & 2972                                                     & 84.6\%                                                   & 85.0\%                                                    & 83.8\%                                                   & 81.7\%                                                       & 82.6\%                                                      & 0.0189                                                & 72.0                                                     & 1183                                                \\
MobileNet-GraspNeXt    & 2712                                                     & 84.3\%                                                   & 84.6\%                                                    & 83.7\%                                                   & 80.7\%                                                       & 81.2\%                                                      & 0.0104                                                & 70.9                                                     & 1189                                                \\
Fast GraspNeXt         & \textbf{1106}                                            & \textbf{87.9\%}                                          & \textbf{85.4\%}                                           & \textbf{85.0\%}                                          & \textbf{85.1\%}                                              & \textbf{84.6\%}                                             & \textbf{0.0095}                                       & \textbf{17.8}                                            & \textbf{259}                                        \\ \bottomrule
\end{tabular}
}
\end{table*}

\begin{figure*}
    \centering
    \includegraphics[width=0.24\linewidth]{figures/suction_heatmap.jpg}
    \hfill
    \includegraphics[width=0.24\linewidth]{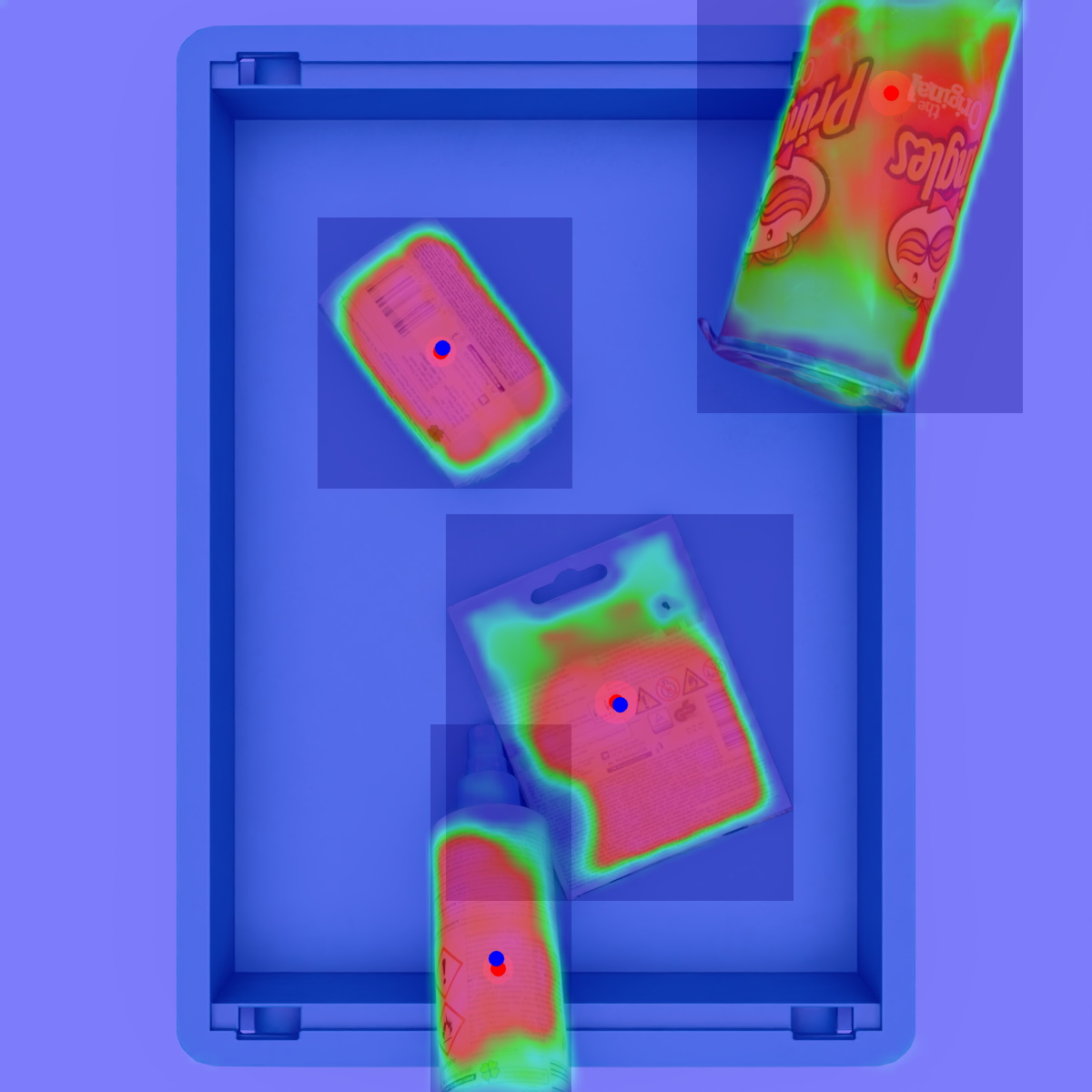}
    \hfill
    \includegraphics[width=0.24\linewidth]{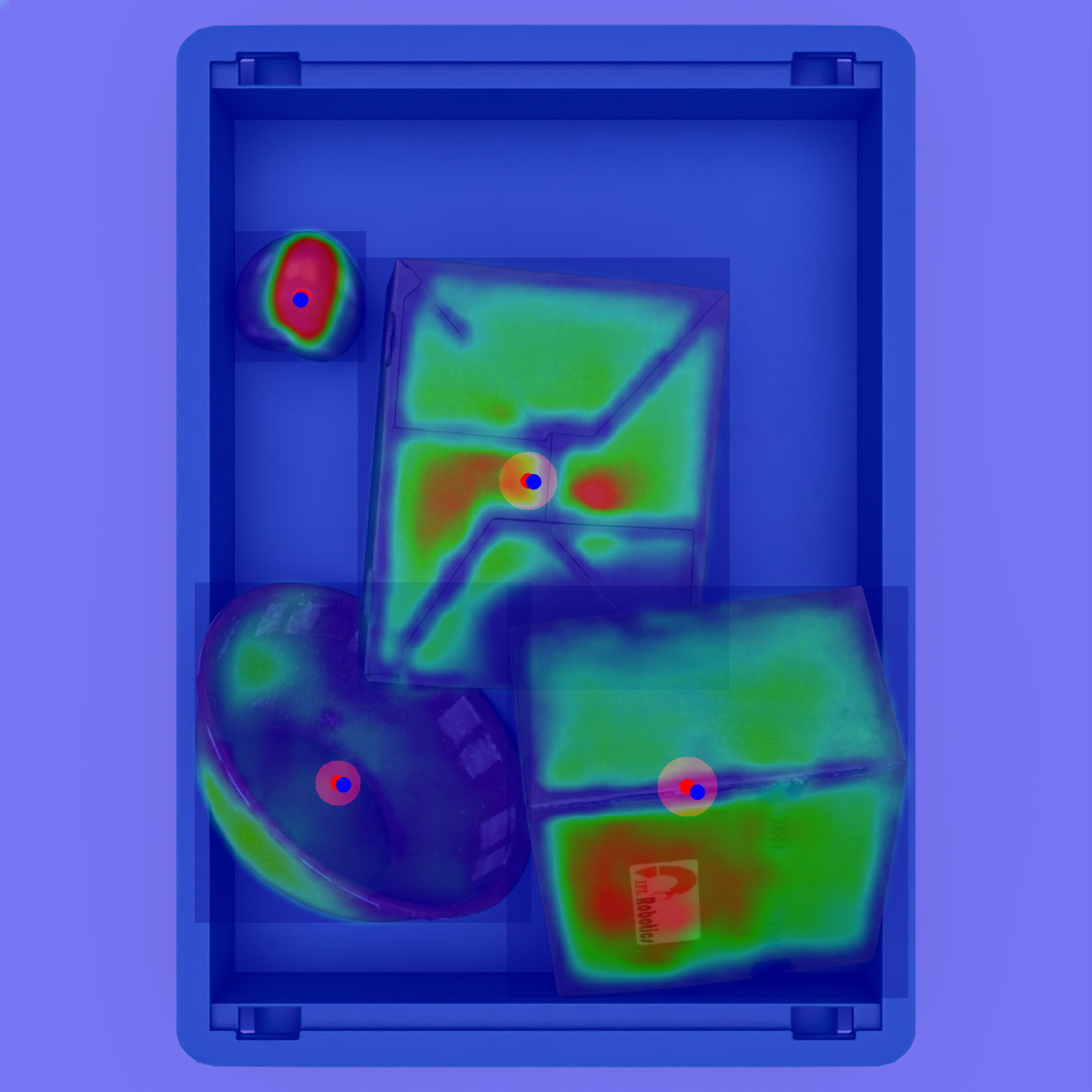}
    \hfill
    \includegraphics[width=0.24\linewidth]{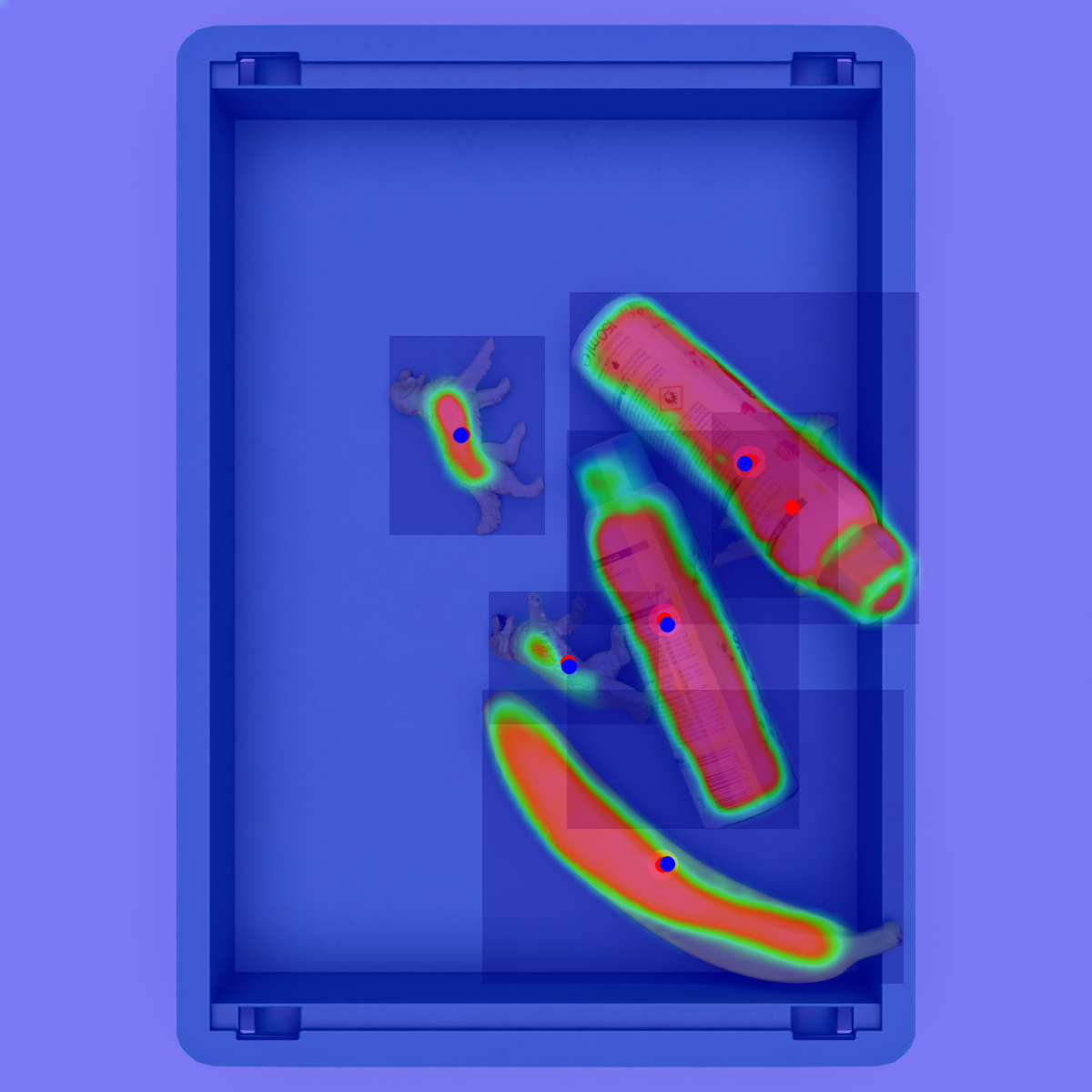} \\
    \vspace{0.1in}
    \includegraphics[width=0.24\linewidth]{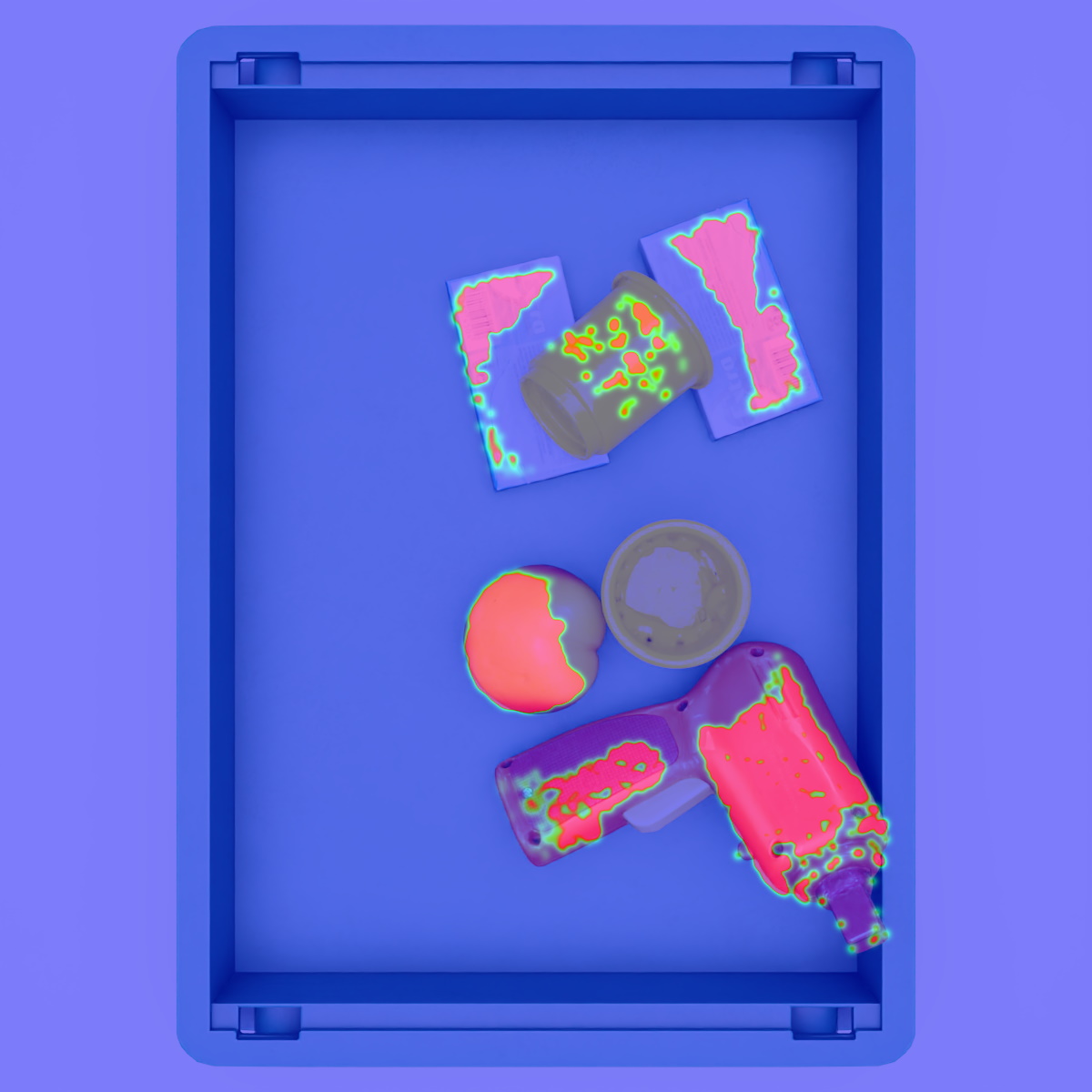}
    \hfill
    \includegraphics[width=0.24\linewidth]{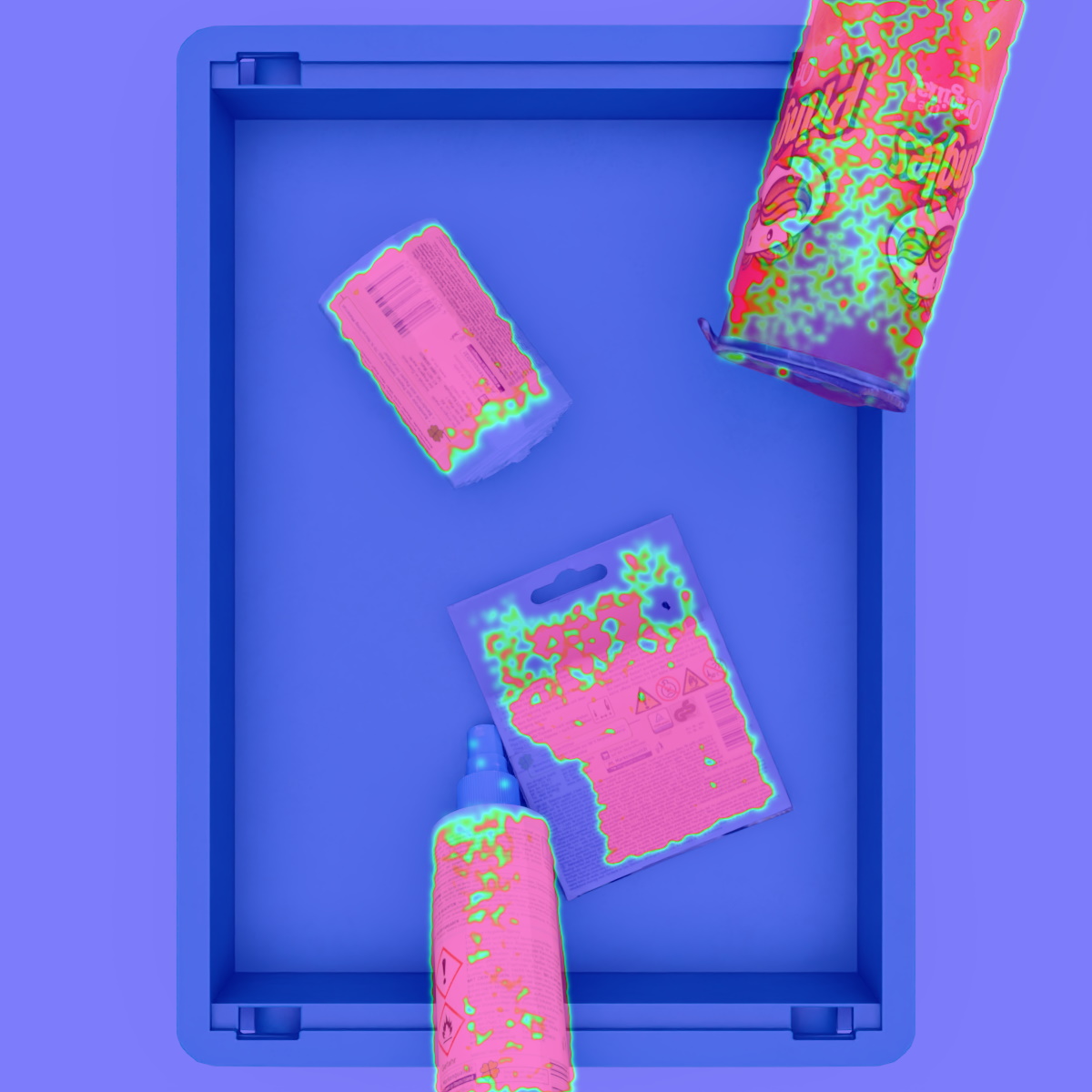}
    \hfill
    \includegraphics[width=0.24\linewidth]{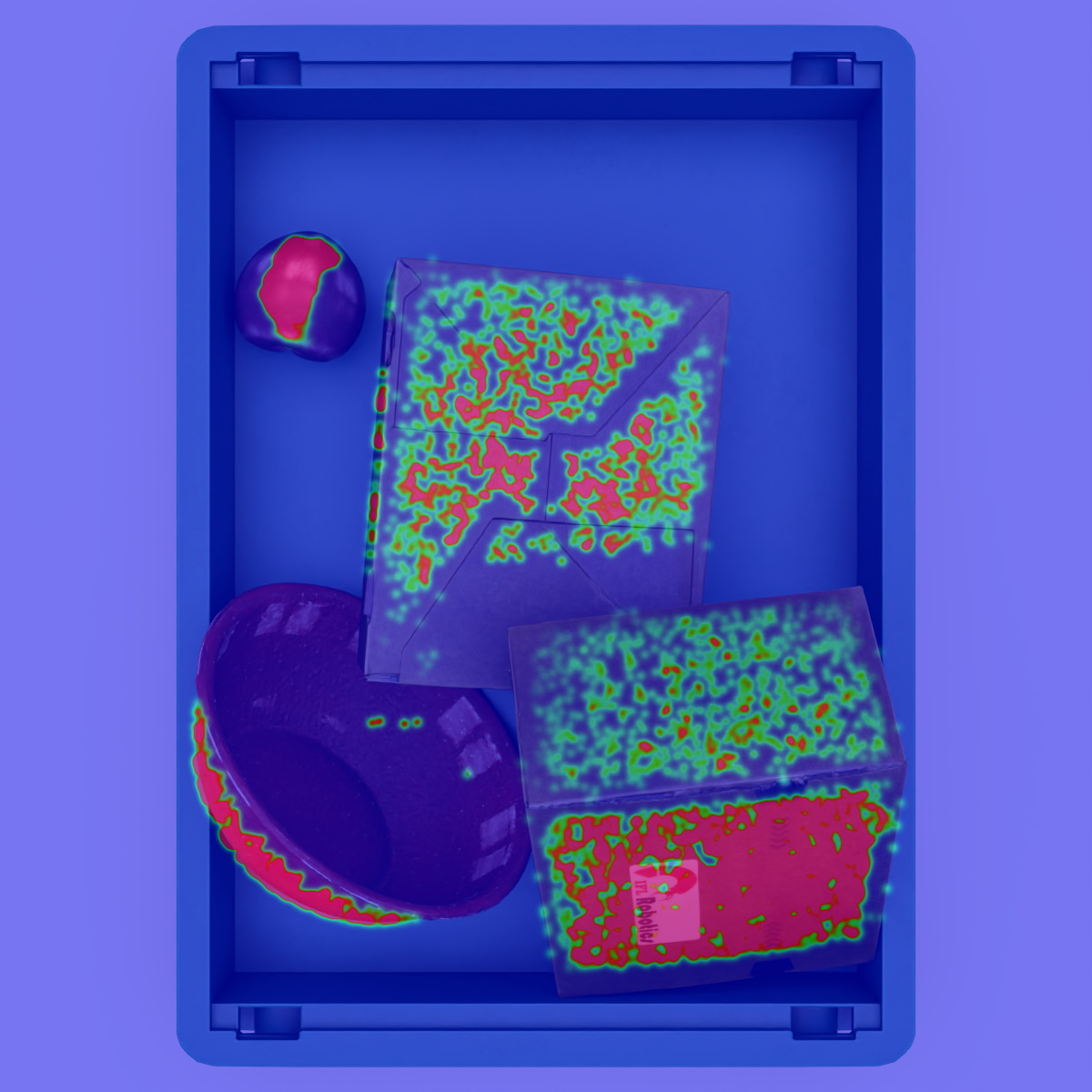}
    \hfill
    \includegraphics[width=0.24\linewidth]{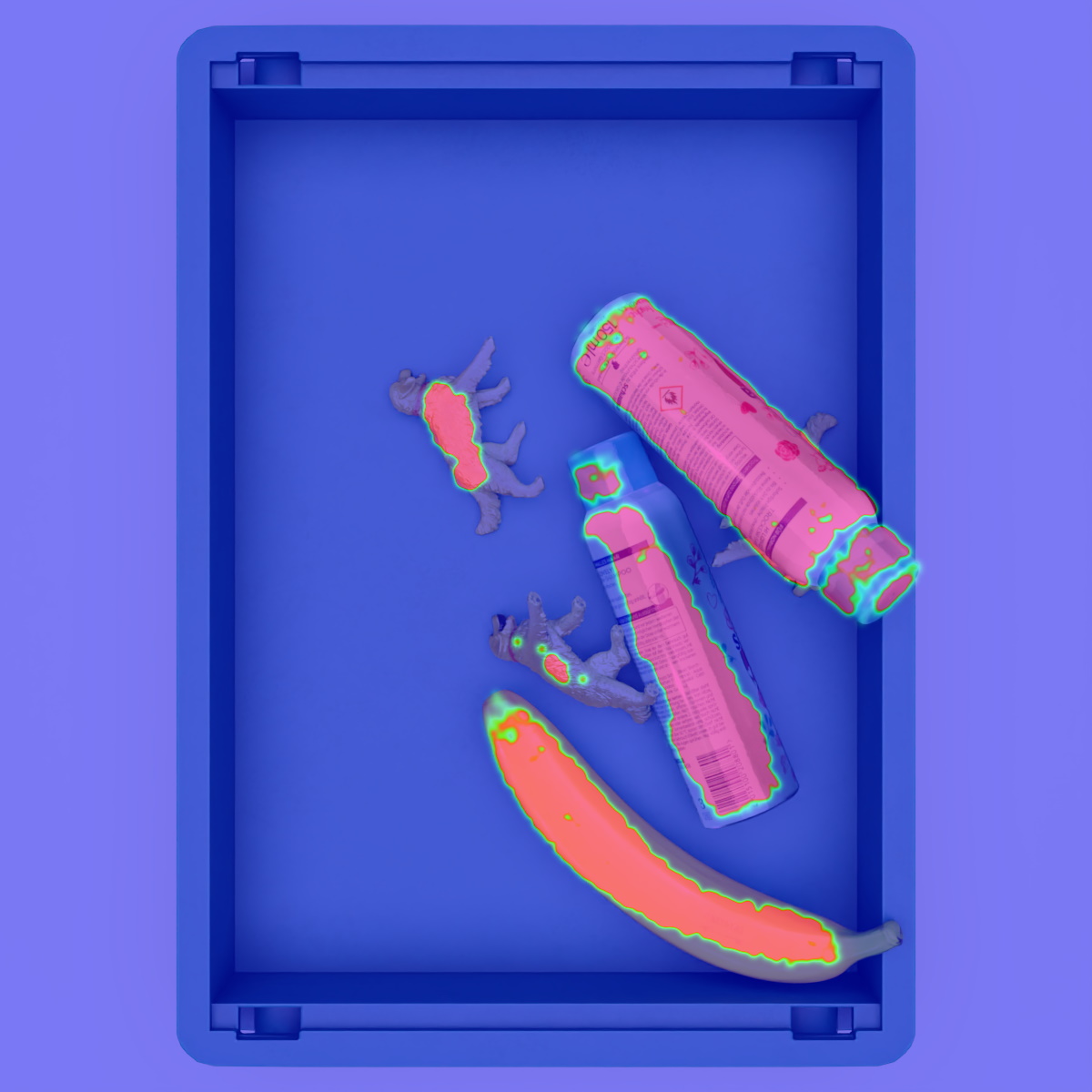}
    \caption{(top) Predicted suction grasp heatmaps produced by the proposed Fast GraspNeXt. (bottom) Example ground truth suction grasp heatmaps.}
    \label{fig:example2}
\end{figure*}

\subsection{Dataset}

We evaluate the performance of the proposed Fast GraspNeXt on the MetaGraspNet~\cite{gilles2022metagraspnet} benchmark dataset to explore the efficacy. This large-scale robotic grasping benchmark dataset contains 217k images across 5884 scenes featuring 82 different objects. We use 60\%, 20\%, and 20\% of the scenes for training, validation, and testing respectively. Average precision (AP) evaluation was conducted for amodal object bounding box, visible object mask, amodal object mask, and object center of mass. Occlusion accuracy evaluation was conducted to evaluate occlusion predictions, while mean squared error (MSE) evaluation was conducted to evaluate suction grasp heatmap predictions.  Our experiments use the class agnostic labels which put all objects into one class category, so that it can be readily deployed in industrial scenarios with novel, unseen items.

\subsection{Training and Testing Setup}
In addition to the proposed Fast GraspNeXt, we evaluated the performance of efficient multi-task network designs leveraging ResNet-50~\cite{he2016deep}, EfficientNet-B0~\cite{tan2019efficientnet}, and MobileNetV3-Large~\cite{howard2019searching} as backbones paired with our multi-task network architecture design but without utilizing the generative network architecture search strategy.  Both EfficientNet and MobileNetV3 are widely-used, state-of-the-art efficient backbones, making them well-suited for this comparison. Those network architectures are designated as ResNet-GraspNeXt, EfficientNet-GraspNeXt, and MobileNet-GraspNeXt, respectively. 

For training, we use a base learning rate of 0.03, SGD optimizer with momentum of 0.9, and weight decay of 0.0001 for all experiments. Learning rate step decay are performed at 67\% and 92\% of the total epochs with gamma of 0.1.  All network architectures are trained with the full image size of 1200$\times$1200 pixels with batch size of 2.  Empirical results found that the above training strategy yielded the best performance for all tested architectures. 

Inference time evaluations are executed with batch size of 1 to reflect the robotic grasping environment which prioritise lowest possible inference latency instead of potential speed benefit of batched inference. We evaluate the inference time on the NVIDIA Jetson TX2 embedded processor with 8 GB of memory, which is widely used for embedded robotics applications in manufacturing and warehouse scenarios.

\subsection{Results and Analysis}
\cref{tab:results} shows the quantitative performance results and model complexity of the proposed Fast GraspNeXt compared to ResNet-GraspNeXt, EfficientNet-GraspNeXt, and MobileNet-GraspNeXt. We can observe that leveraging state-of-the-art efficient backbone architectures EfficientNet-B0 and MobileNetV3-Large enables noticeably faster inference time and lower architectural complexity when compared to leveraging ResNet-50 but results in a noticeable drop in amodal bbox AP and amodal mask AP performance.  In contrast, the proposed Fast GraspNeXt is $>$3.15$\times$, $>$2.68$\times$, and $>$2.45$\times$ faster on the Jetson TX2 embedded processor compared to ResNet-GraspNeXt, EfficientNet-GraspNeXt, and MobileNet-GraspNeXt, respectively, while improves the performance across all test tasks. Specifically, Fast GraspNeXt improves the amodal bbox AP, visible mask AP, amodal mask AP, occlusion accuracy, center of mass AP, and averaged heatmap MSE by 2.9\%, 0.4\%, 0.6\%, 3.4\%, 2.0\%, and 8.7\% respectively compared to the second best results. 

In terms of architectural complexity, Fast GraspNeXt is 5.2$\times$ smaller then ResNet-GraspNeXt which has the second best amodal bbox AP and amodal mask AP, 4$\times$ smaller then EfficientNet-GraspNeXt which has the second best visible mask AP and center of mass AP, and 4$\times$ smaller then MobileNet-GraspNeXt. In terms of computational complexity, Fast GraspNeXt is 5.1$\times$, 4.6$\times$, and 4.6$\times$ lower FLOPs than ResNet-GraspNeXt, EfficientNet-GraspNeXt, and MobileNet-GraspNeXt respectively.  Example ground truth suction grasp heatmaps along with the predicted suction grasp heatmaps produced by proposed Fast GraspNeXt are shown in \cref{fig:example2}. 

As such, the above experimental results demonstrated that the proposed Fast GraspNeXt achieves significantly lower architectural complexity and computational complexity while possessing improved AP across test tasks compared to designs based on state-of-the-art efficient architectures. Furthermore, these experiments demonstrated that Fast GraspNeXt achieves significantly faster inference time on the NVIDIA Jetson TX2 embedded processor, making it well-suited for robotic grasping on embedded devices in real-world manufacturing environments.  Future work involves exploring this generative approach to network architecture search for other embedded robotics applications in manufacturing and warehouse scenarios.

\subsubsection*{Acknowledgements}
This work was supported by the National
Research Council Canada (NRC) and German Federal Ministry for Economic Affairs
and Climate Action (BMWK) under grant 01MJ21007B. 

{\small
\bibliographystyle{ieee_fullname}
\bibliography{manuscript}
}

\end{document}